\documentclass[letterpaper, 10 pt, conference]{ieeeconf}
\IEEEoverridecommandlockouts
\usepackage{graphicx}
\usepackage{caption}
\usepackage{subcaption}
\usepackage{multirow}
\usepackage{tabularx}
\usepackage{lipsum}
\usepackage{caption}
\usepackage{amsmath,amssymb}
\usepackage[dvipsnames,table,xcdraw, HTML]{xcolor}
\usepackage{hyperref}
\hypersetup{
    colorlinks=true,
    linkcolor=orange,
    filecolor=magenta,      
    urlcolor=orange,
    citecolor=orange,
}
\usepackage[capitalise, nameinlink]{cleveref}
\makeatletter
\let\NAT@parse\undefined
\makeatother
\usepackage[numbers,sort,compress]{natbib}
\usepackage{hyperref}
\hypersetup{
    colorlinks=true,  
    linkcolor=red,    
    filecolor=magenta, 
    urlcolor=blue     
}

\usepackage{color, colortbl, soul}
\usepackage{multicol}
\usepackage{listings} 
\usepackage{tcolorbox}
\let\labelindent\relax
\usepackage{enumitem}
\usepackage{booktabs}
\usepackage{siunitx}
\usepackage{amsfonts} 

\usepackage{algorithm}
\usepackage{algpseudocode}
\usepackage[ruled,vlined,linesnumbered,algo2e]{algorithm2e}

\usepackage{xcolor}
\usepackage{soul}  
\usepackage{color, soul}
\usepackage[customcolors]{hf-tikz}
\usepackage{dsfont} 
\usepackage{colortbl}
\usepackage{multirow}
\usepackage{transparent}

\definecolor{lightgreen}{rgb}{0.8,1,0.8}

\definecolor{perfblue}{RGB}{64, 114, 175}

\DeclareSymbolFont{rsfs}{U}{rsfs}{m}{n}
\DeclareSymbolFontAlphabet{\mathscrsfs}{rsfs}
\DeclareMathAlphabet{\mathcal}{OMS}{cmsy}{m}{n}

\newcommand{\methodname}{{\textsc{X-Diffusion}}\xspace}

\title{\LARGE \bf \methodname: Training Diffusion Policies on\\Cross-Embodiment Human Demonstrations}
\author{
    Maximus A. Pace$^*$ \quad
    Prithwish Dan$^*$ \quad
    Chuanruo Ning \quad
    Atiksh Bhardwaj \quad
    Audrey Du \quad \\
    Edward W. Duan \quad
    Wei-Chiu Ma$^\dagger$ \quad
    Kushal Kedia$^\dagger$ \\
    \normalfont{Cornell University}\\
    \url{https://portal-cornell.github.io/X-Diffusion/}
}
\begin{document}

\renewcommand{\footnoterule}{%
  \kern -3pt
  \hrule width 0.5\columnwidth
  \kern 2.6pt
}
\twocolumn[{%
\renewcommand\twocolumn[1][]{#1}%
\maketitle
\begin{center}
\captionsetup{type=figure}
\includegraphics[width=1.0\textwidth]{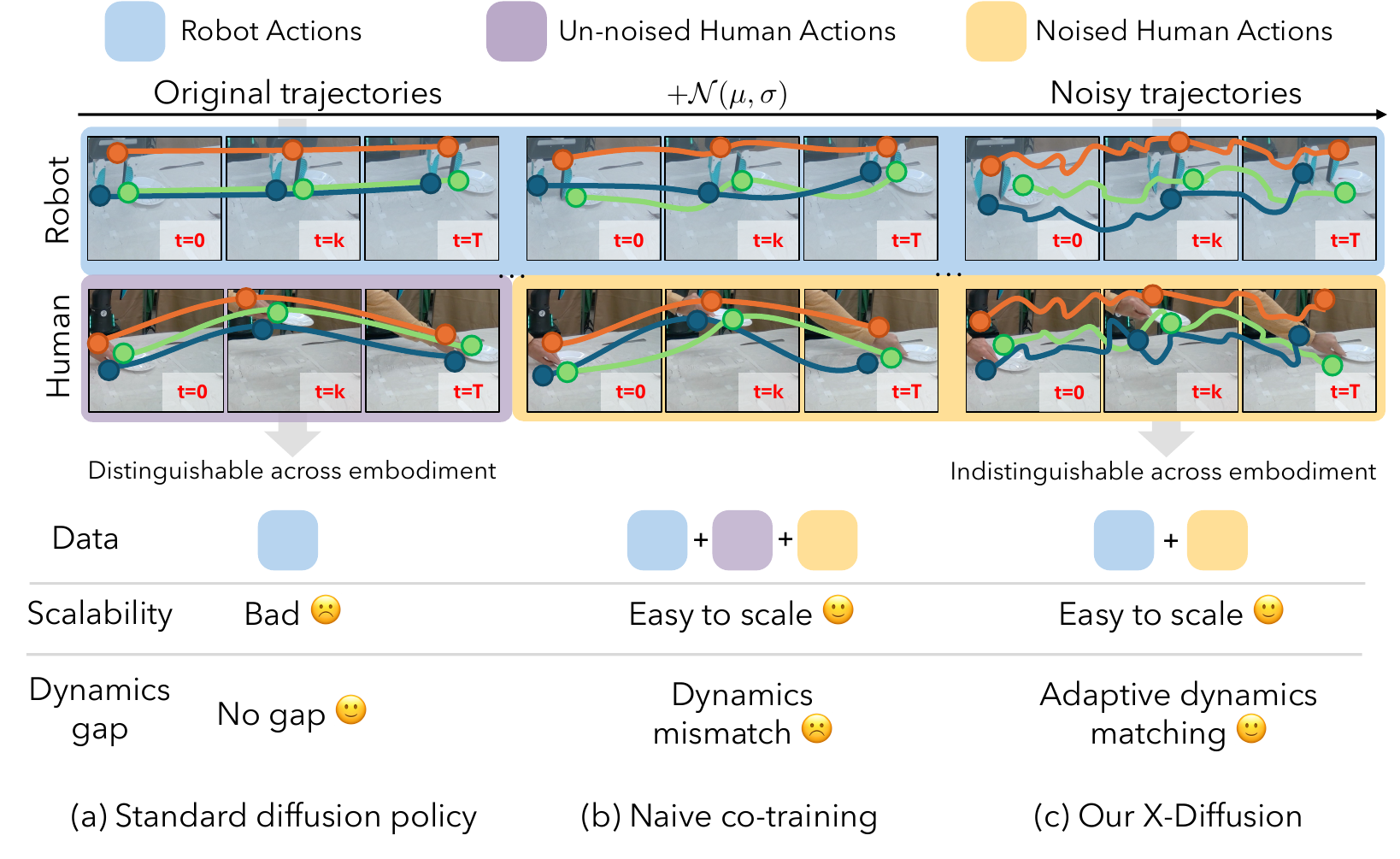}
 \captionof{figure}{\textbf{Overview of \methodname:} We introduce \methodname, a cross-embodiment learning framework that trains Diffusion Policies on human demonstrations even when their actions are not directly executable by the robot. Prior methods typically co-train on mixed human and robot datasets, which often causes the policy to learn actions that are dynamically infeasible on the robot. Instead, \methodname integrates human actions into Diffusion Policy training only when they are sufficiently noised in the forward diffusion process, such that they are indistinguishable from robot actions. This enables the utilization of broad human data without sacrificing dynamic feasibility on the robot.
 }
    \label{fig:introduction}
\end{center}
}]

\let\thefootnote\relax\footnotetext{$^*$ Equal contribution.\quad $^\dagger$ Equal advising.}

\thispagestyle{empty}
\pagestyle{empty}

\begin{abstract}
Human videos are a scalable source of training data for robot learning. However, humans and robots significantly differ in embodiment, making many human actions infeasible for direct execution on a robot. Still, these demonstrations convey rich object-interaction cues and task intent. Our goal is to learn from this coarse guidance without transferring embodiment-specific, infeasible execution strategies. Recent advances in generative modeling tackle a related problem of learning from low-quality data. In particular, Ambient Diffusion is a recent method for diffusion modeling that incorporates low-quality data only at high-noise timesteps of the forward diffusion process. Our key insight is to view human actions as noisy counterparts of robot actions. As noise increases along the forward diffusion process, embodiment-specific differences fade away while task-relevant guidance is preserved. Based on these observations, we present \methodname, a cross-embodiment learning framework based on Ambient Diffusion that selectively trains diffusion policies on noised human actions. This enables effective use of easy-to-collect human videos without sacrificing robot feasibility. Across five real-world manipulation tasks, we show that \methodname improves average success rates by 16\% over naive co-training and manual data filtering.
\end{abstract}

\section{Introduction}

Imitation learning (IL) is an effective and flexible method for teaching robot skills, but collecting large amounts of robot data is costly and slow. Human video demonstrations offer a scalable alternative, since they are easier and faster to collect. However, such data cannot be directly used to train state-of-the-art IL methods~\cite{chi2024diffusionpolicy, Zhao2023LearningFB} because humans and robots significantly differ in embodiment.

To partially address this challenge, recent works propose to map human motions into the robot's action space~\cite{Ren2025MotionTA, Haldar2025PointPU, Lepert2025PhantomTR}. By utilizing advances in 3D hand-pose estimation~\cite{pavlakos2024reconstructing}, hand motions extracted from human videos can be converted into robot end-effector actions via kinematic retargeting, making it possible to learn from large-scale human video datasets ~\cite{Tao2025DexWildDH, Liu2025EgoZeroRL,Lepert2025MasqueradeLF, Shi2025ZeroMimicDR}. 
Yet such mappings only unify the representation of actions, not their physical realizability. Human executions often involve dynamics and contact strategies that are fundamentally mismatched with the robot's embodiment.

Consider the example in Fig.~\ref{fig:introduction}. Even for a simple manipulation task, humans and robots differ in execution style. When moving the plate, a human can dexterously slide their fingers underneath to pick it up, whereas a robot with a parallel-jaw gripper may more reliably push or slide the plate across the surface. 
This naturally raises a key question: how should we treat these human demonstrations? 
Even when the execution itself is not robot-feasible, human motions still provide rich cues about how objects could be manipulated and interacted with. Should we ignore the potential feasibility gap and train on all human data indiscriminately, or should those misaligned with the robot’s capabilities be identified and discarded to prevent degrading policy performance?

Similar challenges exist in the field of generative modeling, where naively training on a mixture of low-quality and high-quality data often degrades model performance~\cite{zhou2023lima, xia2024less}. While prior works filter low-quality samples~\cite{wang2023openchat, li2024superfiltering} or extract signals from noisy or corrupted data~\cite{bora2018ambientgan,lehtinen2018noise2noise,daras2025ambient, kondylatos2025probabilistic}, Ambient Diffusion~\cite{daras2023ambient, daras2025ambientdiffusionomnitraining} offers an exciting alternative by strategically integrating low-quality data into higher-noise timesteps of diffusion.
In this paper, we build upon recent progress in learning from noisy data \cite{daras2023ambient, daras2023consistent, daras2024consistent, daras2024much, daras2025ambientdiffusionomnitraining} to advance cross-embodiment learning. We show how these ideas can be integrated into prevailing robot-learning frameworks \cite{chi2024diffusionpolicy}.

Our key idea is to \emph{view human actions as a noisy counterpart to robot actions}. After mapping human and robot trajectories into a shared action space, embodiment-specific dynamics mismatches can be interpreted as manifestations of noise.
During training, Diffusion Policies learn denoising networks by adding noise to action data. When a sufficient amount of noise is applied to both human and robot actions, low-level embodiment differences fade away while preserving the underlying task structure. Consequently, selectively training Diffusion Policies on noised human actions improves task performance without sacrificing robot feasibility.

Towards this goal, we train a classifier to distinguish between noised human and robot actions in the forward diffusion process. We then define the \textit{minimum indistinguishability step} as the earliest diffusion step where the classifier can no longer discern an action's source embodiment. Actions that are compatible with robot kinematics and dynamics are integrated at lower noise levels, while actions that diverge from the robot’s execution style are only included at higher noise levels. As a result, feasible human and robot demonstrations provide precise, low-level supervision throughout the diffusion process, whereas mismatched human actions contribute only coarse, high-level guidance. This enables Diffusion Policies to extract useful signals from all human data while avoiding degradation from execution mismatches.

We validate \methodname on five real-world manipulation tasks exhibiting varying human-robot execution mismatch. While prior approaches that naively co-train on human data may generate infeasible robot actions, selectively training on human actions at high-noise levels improves upon naive co-training and even surpasses manual data filtering. \methodname outperforms a range of cross-embodiment learning baselines by an average of 16\% in task success.

\section{Related Work}
Our work is related to the following topics:

\textbf{Learning from Human Hand Motion.}
Advances in hand-pose estimation have enabled retargeting actionless human videos into robot actions. One approach is to track 6DoF hand trajectories and map them to the robot end-effector~\cite{Bharadhwaj2023ZeroShotRM, wang2023mimicplay}. Other works define corresponding keypoints between humans and robots to unify their data representations~\cite{Ren2025MotionTA, Haldar2025PointPU}, overlaying rendered robot arms on human videos~\cite{Lepert2025SHADOWLS, Lepert2025PhantomTR, bahl2022human}. Open-world vision models have further enabled object-aware retargeting~\cite{zhu2024vision, Vitiello2023OneShotIL, Li2024OKAMITH}. These methods assume that retargeted hand motions will transfer cleanly to the robot, but this often fails in practice due to embodiment mismatch.


\textbf{Extracting Rewards from Human Data.}
Reinforcement learning (RL) approaches leverage human data by defining rewards from tracking reference motion~\cite{2018-TOG-deepMimic, Yuan2025HERMESHE}, object-centric signals in real-to-sim-to-real pipelines~\cite{dan2025xsim, Ga2025CrossingTH}, and classifier judgments of task success~\cite{schmeckpeper2020learning}. However, these approaches are limited by the requirement of a realistic simulator or costly and unsafe real-world interactions. In contrast, we train Diffusion Policies directly on mixed human–robot data without requiring environment interactions.
\begin{figure*}[t!]
    \centering
    \includegraphics[width=\textwidth]{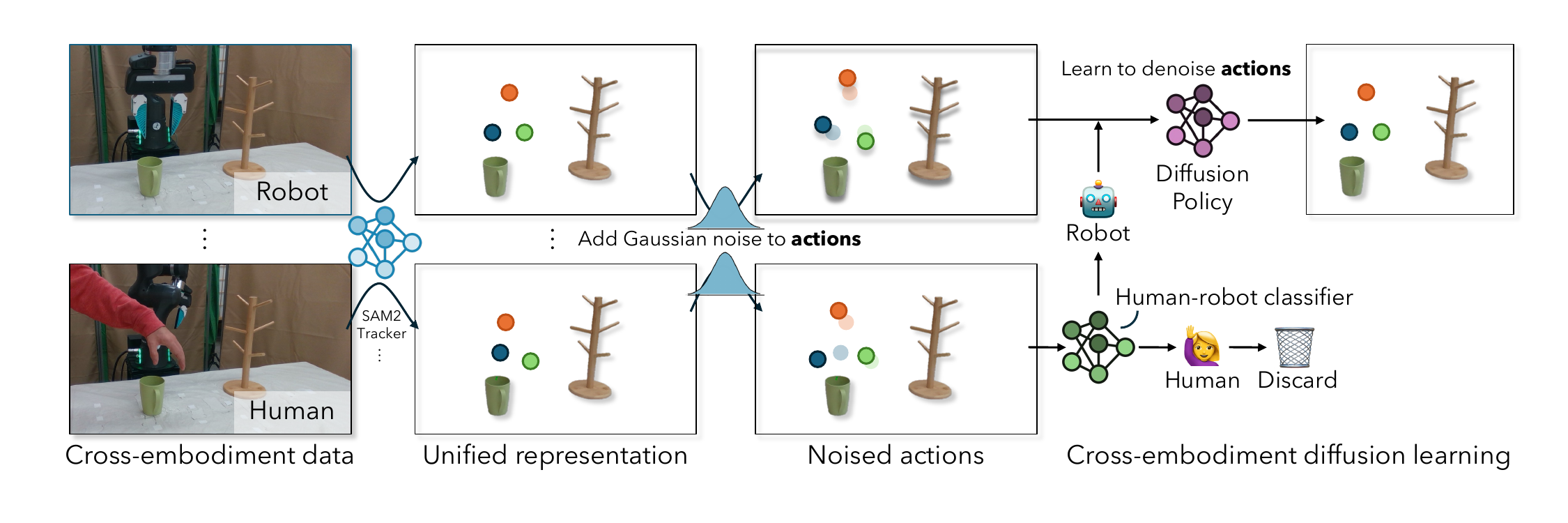}
    \captionsetup{width=\textwidth}
    \vspace{-9mm}
    \caption{\textbf{Pipeline:} \methodname first unifies the state and action representation. State is represented by a colored segmentation mask of relevant objects using Grounded SAM 2~\cite{ravi2024sam2segmentimages, ren2024grounded}. Action is represented via end-effector/human hand pose utilizing HaMeR~\cite{pavlakos2024reconstructing} for retargeting. To determine if the policy should learn to denoise noisy human actions, \methodname utilizes a classifier trained to distinguish the source embodiment of noised actions. Actions are only included for training the denoising process if the classifier is fooled into thinking it's from a robot.}
    \label{fig:pipeline}
    \vspace{-3mm}
\label{fig:realworld-tsne}
\end{figure*}

\textbf{One-Shot Imitation from Human Videos.}  
Prior work has explored one-shot imitation, where robots attempt a task from a single human demonstration. Some methods learn correspondences from paired human–robot videos~\cite{Jang2022BCZZT, Jain2024Vid2RobotEV}, unify visual embeddings of humans and robots~\cite{kedia2024one, xu2023xskill}, use a human video as a guide to retrieve task-relevant behaviors~\cite{shah2025mimicdroid, vosylius2025instantpolicyincontextimitation}, or prompt pretrained policies with retargeted trajectories~\cite{park2025demodiffusiononeshothumanimitation}, but these require costly paired data, large teleoperated datasets, or heavy reliance on base policies. Our method learns directly from multiple human demonstrations.


\textbf{Learning from Sub-Optimal Data.}  
Collecting large amounts of high-quality robot data is prohibitively expensive. As a result, recent work has focused on estimating demonstration quality via costly online interactions~\cite{chen2025curatingdemonstrationsusingonline, agia2025cupidcuratingdatarobot} or proxy loss metrics~\cite{hejna2024remixoptimizingdatamixtures} that often correlate poorly with real-world performance.
In generative modeling, prior works have focused on extracting clean signals from noisy or uncurated datasets~\cite{zhou2023lima, lehtinen2018noise2noise, bora2018ambientgan, dai2023emuenhancingimagegeneration}.
Our method builds upon Ambient Diffusion \cite{daras2023ambient, daras2023consistent, daras2024consistent, daras2024much, daras2025ambientdiffusionomnitraining}, a method for training diffusion models on low-quality data to produce high-quality samples. Its core principle is to incorporate low-quality samples into training only when they have been sufficiently noised in the diffusion process. This enables the diffusion model to learn from large amounts of low-quality data without degrading its outputs. Applying this to cross-embodiment robot learning, we treat dynamically infeasible demonstrations as low-quality data, exploiting Ambient Diffusion to adaptively extract useful guidance from uncurated human demonstrations.

\section{Problem Formulation and Background}
Our goal is to learn a robot policy $\pi_\theta (\mathbf{A_t} | s_t)$, which predicts a sequence of future actions $\mathbf{A_t}=a_{t:t+S}$ over the next $S$ timesteps given the current robot state $s_t$. Training relies on two sources of supervision: a small, high-quality dataset of robot demonstrations \(\mathcal{D}_R\) and a larger dataset of human demonstrations \(\mathcal{D}_H\). Each dataset contains trajectories of state--action pairs \(\xi = \{s_t, a_t\}_{t=1}^T\).

\textbf{Co-Training of Robot Policies.} Cross-embodiment datasets are typically leveraged for policy learning by \emph{co-training} with the robot dataset. A straightforward approach is to simply combine the robot dataset \(\mathcal{D}_R\) and the human dataset \(\mathcal{D}_H\) and train on the aggregated mixture:
\begin{equation}
\mathcal{L}_{\text{co-train}}(\theta) = \mathbb{E}_{(s_t,\mathbf{A_t}) \sim \mathcal{D}_R \cup \mathcal{D}_H} \left[ \ell\big(\pi_\theta(s_t), \mathbf{A_t}\big) \right],
\label{eq:cotrain}
\end{equation}
where \(\ell\) is the behavior cloning loss. This assumes human and robot data have interchangeable dynamics, i.e., $p_H(\mathbf{A_t}=a_{t:t+S}|s_t) \approx p_R(\mathbf{A_t}=a_{t:t+S}|s_t)$. However, differences in embodiment and execution style mean that human actions are often physically infeasible for the robot. As a result, naive co-training can significantly degrade policy performance, motivating the need for more selective co-training strategies.

\textbf{Ambient Diffusion.} Ambient Diffusion~\cite{daras2023ambient, daras2023consistent, daras2024consistent, daras2025ambientdiffusionomnitraining, daras2025ambient} is a recent method that trains diffusion models on low-quality data under sufficient noise. Their key insight is that high- and low-quality distributions $p_{\rm high}$ and $p_{\rm low}$ are close ($\epsilon$-merged~\cite{daras2025ambient}) after $k$ steps in the forward diffusion process if $D_{KL}\!\left(p_{\rm low}^k \;\|\; p_{\rm high}^k\right) \leq \epsilon$, enabling the use of low-quality data in high-noise regimes. We connect this idea to robot policy learning: when training Diffusion Policies~\cite{chi2024diffusionpolicy}, we view human and robot demonstrations as low- and high-quality samples, respectively, learning from noised human actions only when they match the robot's dynamics.

\textbf{Unifying State and Action Spaces.}
Following prior work~\cite{Ren2025MotionTA, Haldar2025PointPU}, we unify the cross-embodiment data into a shared state $s_t=(q_t,o_t)$ and action $a_t=q_{t+1}$. The proprioception $q_t\in\mathbb R^7$ contains the end-effector 3D position, rotation, and gripper state. For human data, we assume access to the following: (i) single-hand demonstrations that begin with an open grasp, and (ii) two calibrated RGB cameras. Using HaMeR~\cite{pavlakos2024reconstructing}, we detect 2D hand keypoints in each view and triangulate them to the 3D robot frame. The grasp point is the mean of the thumb and index fingertips; orientation is obtained by fitting a local hand frame and retargeting to the robot end-effector following prior work~\cite{Ren2025MotionTA, Haldar2025PointPU}. Gripper state is inferred using the distance between the thumb and index keypoints. To reduce the visual domain gap, we segment task-relevant objects with Grounded SAM~2~\cite{ravi2024sam2segmentimages, ren2024grounded} and overlay a keypoint rendering of the end-effector pose on each frame, as depicted in Fig.~\ref{fig:pipeline}. The policy input concatenates this masked image with the proprioceptive information.
\section{Approach}\label{sec:approach}
\begin{figure*}[t!]
    \centering
    \includegraphics[width=\textwidth]{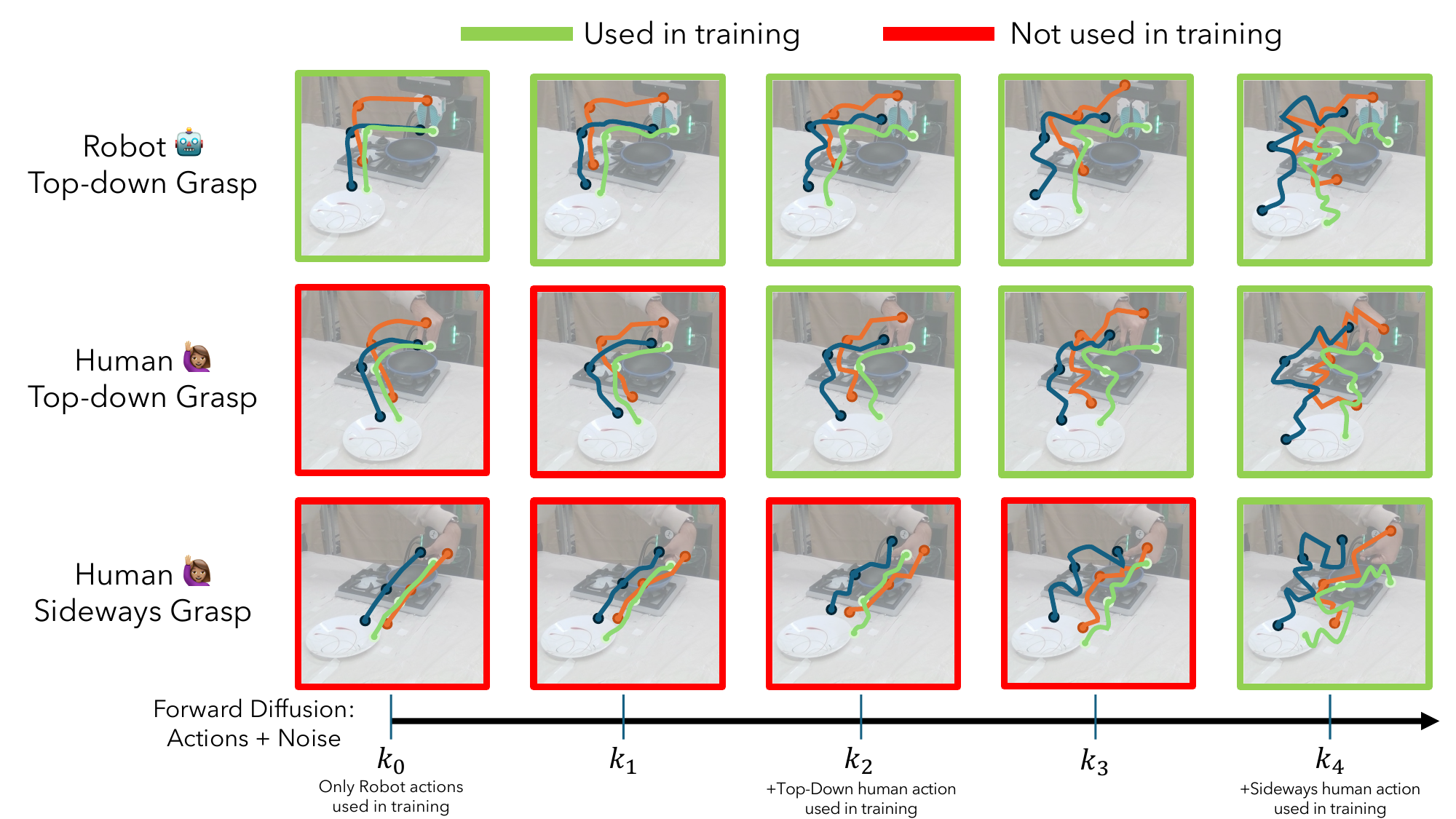}
    \captionsetup{width=\textwidth}
    \vspace{-2mm}
    \caption{\textbf{Visualizing Actions under Noise and Classifier Predictions at various Diffusion Steps}. Humans execute tasks in various ways. For example, when picking and placing a pan, a human can either execute a top-down grasp or a side grasp. Human actions that are feasible for robots (e.g. top-down grasp) overlap with robot action distribution under low noise timesteps. This data fools the classifier into believing it could have been executed by a robot, so we include it in the diffusion denoising process during policy training. In contrast, human actions that are kinematically and dynamically infeasible for robots (e.g. side grasp) are accurately identified as human actions by the classifier until significantly more noise is added in the forward diffusion process, restricting their impact on policy learning to only supervise coarse guidance at high noise. 
    }
    \vspace{-3mm}
\label{fig:classifier}
\end{figure*}
Naive co-training on human and robot demonstrations can degrade performance when execution styles are mismatched. In this section, we present \methodname, a cross-embodiment learning framework based on Ambient Diffusion~\cite{daras2023ambient} to maximally utilize cross-embodiment data for Diffusion Policy learning without degrading performance. \methodname\ first trains a classifier to distinguish between noised human and robot actions. Noised human actions are integrated into policy training only when the classifier is confused about its embodiment. This approach allows us to utilize large datasets of cross-embodiment demonstrations without learning dynamically infeasible robot actions.

\subsection{Cross-Embodiment Equivalence under Noise}
Due to embodiment differences, kinematic retargeting of human hand actions may result in physically infeasible robot motion. Still, human demonstrations provide rich cues for what steps to follow, which objects to interact with, and how to interact with them. The usefulness of these cues depends on their alignment with the robot's action dynamics. 

Diffusion Policies~\cite{chi2024diffusionpolicy} learn by denoising action sequences corrupted with Gaussian noise. Given the clean robot or human action sequence $\mathbf{A_t^0}$, the \emph{forward diffusion process} $q$ produces progressively noisier versions $\mathbf{A_t^1}, \dots, \mathbf{A_t^K}$ via:
\[q(\mathbf{A_t^{k+1}} \mid \mathbf{A_t^k}) = \mathcal{N}\!\left(\sqrt{1-\beta_k}\,\mathbf{A}_t^k,\; \beta_k I\right),\]
 where $\beta_k$ controls the amount of additive Gaussian noise at diffusion step $k$. Our key observation is that the \emph{forward diffusion} process progressively removes embodiment-specific features from actions. As shown in Fig.~\ref{fig:introduction}, \emph{at high noise levels, human and robot trajectories become indistinguishable}. 

Formally, let $p_H^k$ and $p_R^k$ denote the distributions of human and robot actions at diffusion step $k$. Similar to the $\epsilon$-merging time in Ambient Proteins~\cite{daras2025ambient}, we define the \textbf{minimum indistinguishability step} $\mathbf{k^\star}$ as the earliest diffusion step where the two distributions overlap such that they cannot be reliably distinguished:
\[
k^\star = \min \Big\{ k \;\big|\; D_{KL}\!\left(p_H^k \;\|\; p_R^k\right) \leq \epsilon \Big\},
\]
where $\epsilon$ is a small threshold. Intuitively, $k^\star$ identifies the point in the noising process at which human actions are sufficiently abstracted to resemble robot actions. Beyond this step ($k \geq k^\star$), human demonstrations can safely supervise robot policy learning without the transfer of infeasible motions.
\begin{figure*}[t!]
    \centering
    \includegraphics[width=\textwidth]{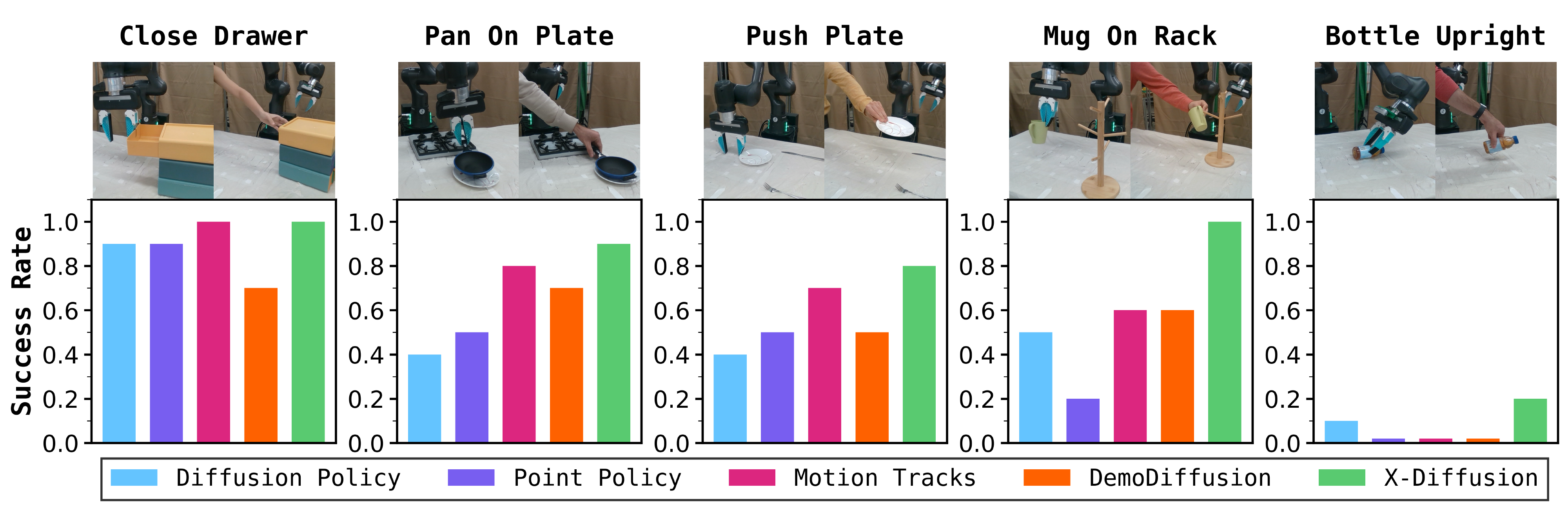}
    \captionsetup{width=\textwidth}
    \vspace{-4mm}
    \caption{\textbf{Performance vs. Baselines:} We report \textit{task success rate} on 5 different manipulation tasks and compare \methodname against a robot-only baseline (Diffusion Policy~\cite{chi2024diffusionpolicy}) and various co-training baselines (Point-Policy~\cite{Haldar2025PointPU}, Motion Tracks~\cite{Ren2025MotionTA}). DemoDiffusion~\cite{park2025demodiffusiononeshothumanimitation} is another diffusion-based method, but it doesn't train the robot policy on human demonstrations. We find that \methodname is the highest performing model on all tasks, effectively incorporating human action data into its training recipe even when execution styles are mismatched. One human and robot demonstration is visualized for each task.}
\label{fig:barplots}
\vspace{-4mm}
\end{figure*}
\subsection{Training a Noised Human-Robot Action Classifier}  
To determine the minimum indistinguishability timestep $k^*$ for each action, we train a classifier that predicts the embodiment of a noised action. This idea is closely related to the classifier used in Ambient Diffusion Omni~\cite{daras2025ambientdiffusionomnitraining} to distinguish between low- and high-quality data. The classifier $c_\theta(\cdot|k, \mathbf{A_t^{k}}, s_t)$ takes in the diffusion step $k$, the noised action sequence $\mathbf{A}_t^{(k)}$, and the current state $s_t$, and outputs the probability of the action originating from the robot ($y=1$) rather than a human ($y=0$). Training samples are drawn from both the human dataset $\mathcal{D}_H$ and robot dataset $\mathcal{D}_R$. Since the human dataset is much larger than the robot dataset $\lvert \mathcal{D}_H \rvert \gg \lvert \mathcal{D}_R \rvert$, we sample actions from each with equal probability to avoid biasing toward the human label. The classifier is optimized with the binary cross-entropy loss:  
\begin{equation}
\begin{aligned}
\mathcal{L}_{\text{class}}(\theta)
=\; &\mathbb{E}_{(k, \mathbf{A_t^k}, s_t)\sim \mathcal{D}_R}\;
\big[-\log c_\theta(k,\mathbf{A_t^k},s_t)\big]
\\[4pt]
+\;&
\mathbb{E}_{(k, \mathbf{A_t^k}, s_t)\sim \mathcal{D}_H}\;
\big[-\log\!\big(1-c_\theta(k,\mathbf{A_t^k},s_t)\big)\big].
\end{aligned}
\label{eq:k*}
\end{equation}
The classifier enables us to annotate human demonstrations with the timestep at which their noised actions become indistinguishable from robot actions. For each human action sequence $\mathbf{A}_t$, we define the {minimum indistinguishability step} $k^\star$ as the earliest diffusion step where the classifier assigns at least 50\% probability to it being a robot action:
\begin{equation}
    k^\star(\mathbf{A}_t) \;=\; 
    \min\left\{ k \;:\; c_\theta(k,\mathbf{A}_t^k,s_t) \;\geq\; 0.5 \right\}.
\end{equation}

\subsection{Classifier Integration into Diffusion Policy}
Diffusion Policies model the reverse process of denoising. Starting from Gaussian noise $\mathbf{A}_t^K$, the reverse model $p_\theta(\mathbf{A}_t^{k-1} \mid k, \mathbf{A}_t^k, s_t)$ iteratively denoises until recovering the clean action sequence $\mathbf{A}_t^0$. Naive co-training (Eq.~\ref{eq:cotrain}) supervises the reverse process using human actions across all diffusion steps. If human data is used indiscriminately at all noise levels, the policy is forced to denoise toward actions that may be kinematically infeasible for the robot.

\textbf{Integration beyond the indistinguishability step.} Our classifier resolves this problem by identifying, for each human action, the minimum indistinguishability step $k^\star$ where the action distribution sufficiently overlaps with the robot action distribution under noise. During Diffusion Policy training, we only integrate human actions into the loss when $k \geq k^\star$ (using Eq.~\ref{eq:k*}). Fig.~\ref{fig:classifier} shows the minimum indistinguishability step on the \texttt{Pan On Plate} task for different human actions. Actions that are kinematically feasible for the robot have low $k^*$ whereas infeasible actions have higher $k^*$. Formally, our Diffusion Policy loss is:
\begin{equation}
\begin{aligned}
\mathcal{L}_{\text{X-DP}}(\theta) =\;&
\mathbb{E}_{(k,\mathbf{A}_t, s_t) \sim \mathcal{D}_R}\;
\ell\!\left(p_\theta, \mathbf{A}_t^k \right) \\[6pt]
+\;&
\mathbb{E}_{(k,\mathbf{A}_t,s_t) \sim \mathcal{D}_H}\;
\mathbf{1}_{\{k \geq k^\star(\mathbf{A}_t)\}}\,
\ell\!\left(p_\theta, \mathbf{A}_t^k \right),
\end{aligned}
\end{equation}

where $\ell$ denotes the denoising loss. This selective integration ensures that we maximally utilize human demonstrations without sacrificing kinematic feasibility of action execution.

\section{Experiments}\label{sec:experiments}
We evaluate the ability of \methodname to learn 5 different manipulation skills from cross-embodiment human data. Our experiments are designed to address four key questions:
\begin{enumerate}[leftmargin=*]
\itemsep0em 
    \item Does \methodname outperform prior cross-embodiment learning approaches?
    \item Does naive co-training generate kinematically or dynamically infeasible motion on the robot?
    \item How does the learned classifier compare to manual data filtering via human annotation?
    \item How does the usefulness of human data vary across tasks?
\end{enumerate}

\textbf{Experimental Setup.} 
For each manipulation task, we collect 5 robot demonstrations and 100 human demonstrations. Human demonstrations are performed with a single hand, while the robot is a 7-DOF Franka Emika Panda arm. We evaluate across five diverse tasks:
\texttt{Close Drawer} (closing a cabinet's top drawer), \texttt{Pan On Plate} (picking a frying pan from a stovetop and placing it on a plate), \texttt{Push Plate} (sliding a plate between a fork and knife), \texttt{Mug On Rack} (inserting a mug's handle onto a rack peg), and \texttt{Bottle Upright} (reorienting a bottle to stand upright).
These tasks span a wide range of manipulation skills and provide a comprehensive benchmark for assessing the value of human data in policy training. We evaluate each method over 10 real-world rollouts per task and report average success rates.

\textbf{Baselines.} We compare against the following baselines:
\begin{enumerate}[leftmargin=*]
\item \textbf{Diffusion Policy~\cite{chi2024diffusionpolicy}}: This method trains only on 5 robot demonstrations, lacking guidance from human data.
\item \textbf{Point Policy~\cite{Haldar2025PointPU}}: This method co-trains a Diffusion Policy on all human and robot data. Its state is object keypoints from DIFT~\cite{tang2023emergent} and Co-Tracker~\cite{karaev2024cotracker} plus hand keypoints.
\item \textbf{Motion Tracks~\cite{Ren2025MotionTA}}: This method co-trains a Diffusion Policy on all human and robot data. It unifies the action space as hand keypoints but uses raw image observations.
\item \textbf{DemoDiffusion~\cite{park2025demodiffusiononeshothumanimitation}}: This method performs the reverse diffusion process using a human policy for the first $60\%$ of steps and a robot policy for the remaining $40\%$.
\end{enumerate}
\subsection{Comparison with Cross-Embodiment Learning Baselines.}
We evaluate \methodname's ability to learn from human demonstrations and compare performance against existing cross-embodiment baselines. We find that \methodname achieves higher success rates across tasks relative to Point Policy, Motion Tracks, and DemoDiffusion (Fig.~\ref{fig:barplots}). Naively co-training on uncurated human demonstrations yields little to no improvements (Motion Tracks, DemoDiffusion) over robot-only training and can even degrade performance (Point Policy) by learning suboptimal robot behaviors.

\begin{figure}[t!]
    \centering
\includegraphics[width=0.48\textwidth]{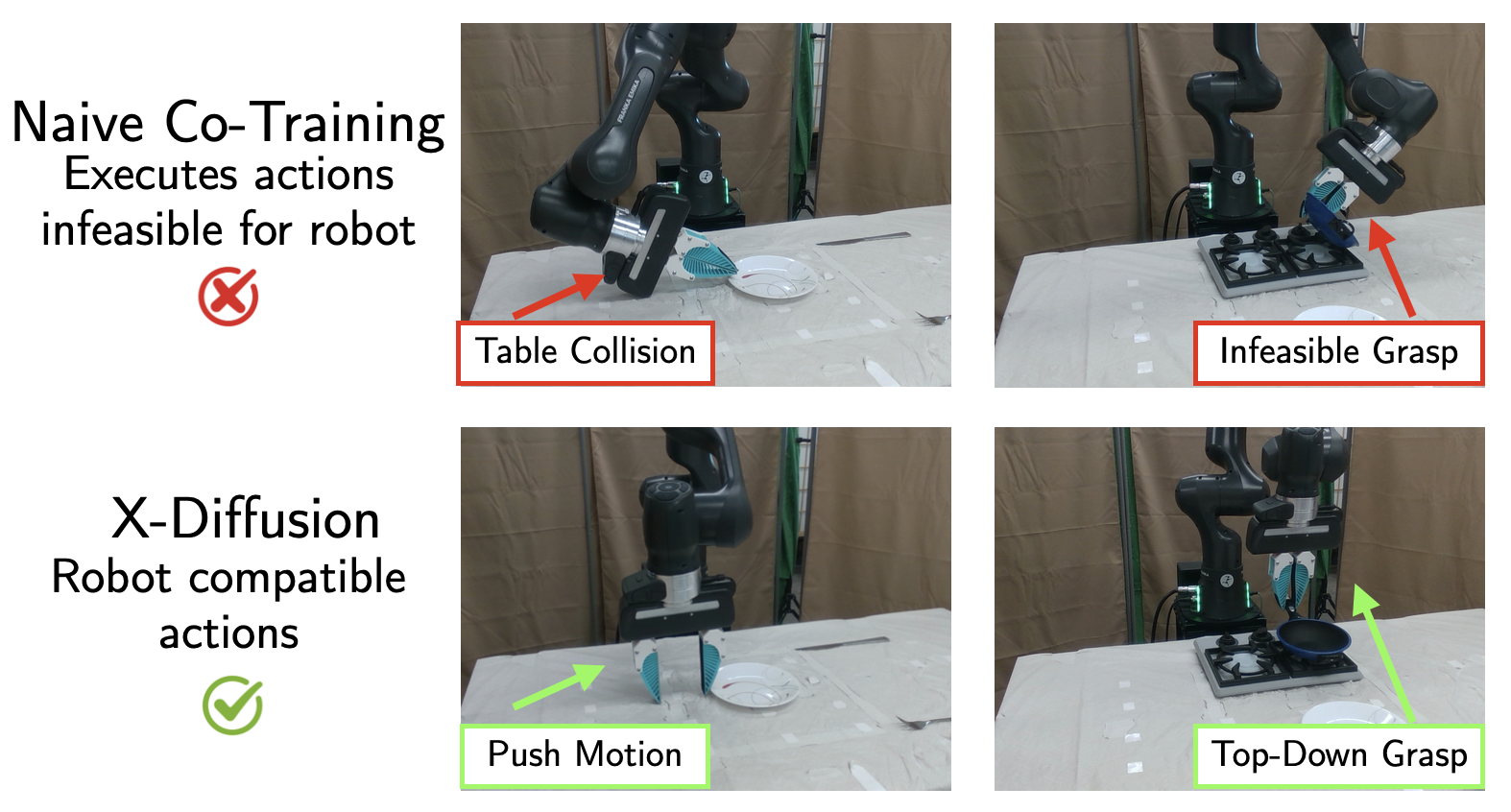}
    \captionsetup{width=0.48\textwidth}
    \vspace{-2mm}
    \caption{\textbf{Naive Co-Training Learns Infeasible Robot Actions:} Including all human data in policy training can incentivize policies to learn strategies demonstrated by humans that are infeasible for robots. On multiple tasks, a human may manipulate objects in ways that are not realizable for a robot.}
\label{fig:infeasible}
\vspace{-6mm}
\end{figure}
Qualitatively, these baselines share a failure mode: executing human actions that are infeasible for the robot (Fig.~\ref{fig:infeasible}). In \texttt{Push Plate} and \texttt{Pan On Plate}, several human demonstrations grasp objects from the side (instead of top-down), a kinematically infeasible strategy for the robot.

Unlike these methods, \methodname leverages its classifier to filter out action sequences that have low probabilities of being classified as robot actions, applying the action denoising loss only to (noisy) human motions indistinguishable from robot motion. This training recipe consistently improves performance over robot-only and naive co-training by carefully including human data from a wider state distribution.

\subsection{Systematic Ablation of Co-Training Data Choices}
To further investigate the human data distribution and its impact on policy learning, we design an experiment with a \textsc{Filtered} policy. We replay human demonstrations on the robot via Inverse Kinematics (IK) and manually filter out unsuccessful trajectories to construct $\mathcal{D}_H^+$, a dataset of feasible human demonstrations. We observe that while nearly all human demonstrations exhibit some degree of mismatch, approximately 50\% of the original demonstrations resulted in kinematic or dynamic failures and were discarded. We train three policies with the same architecture but vary the data:
\begin{itemize}
    \item \textbf{\textsc{Robot Only}:} Trained only on $\mathcal{D}_R$.
    \item \textbf{\textsc{Naive}:} Trained on $\mathcal{D}_R \cup \mathcal{D}_H$.
    \item \textbf{\textsc{Filtered}:} Trained on $\mathcal{D}_R \cup \mathcal{D}_H^+$.
    \item \textbf{\methodname:} Trained on $\mathcal{D}_R \cup \mathcal{D}_H$, discarding human data below the \emph{minimum indistinguishability step} (Sec.~\ref{sec:approach}) during action denoising.
\end{itemize}
Figure~\ref{fig:oracle} shows that \textsc{Filtered} dataset co-training outperforms \textsc{Naive} co-training, confirming the hypothesis that training on infeasible human demonstrations degrades policy performance. \methodname takes an alternate approach---instead of discarding entire trajectories and applying the action denoising loss at all noise levels for successful human trajectories in $\mathcal{D}_H^+$, it adaptively includes human data from $\mathcal{D}_H$ only beyond noise levels where the human and robot data distributions are indistinguishable, thus learning to denoise within the correct distribution for the robot. We visualize this phenomenon in Fig.~\ref{fig:classifier}: as Gaussian noise is added to human actions, our classifier is unable to identify which embodiment executed the actions. We observe that the minimum indistinguishability step is lower for feasible human actions than their infeasible counterparts. \methodname outperforms the \textsc{Filtered} policy across all tasks, demonstrating the ability to extract signal even from infeasible human demonstrations.

\subsection{Quantifying Transfer Learning from Human Data}
\begin{figure}[t]
    \centering
    \includegraphics[width=1.0\columnwidth]{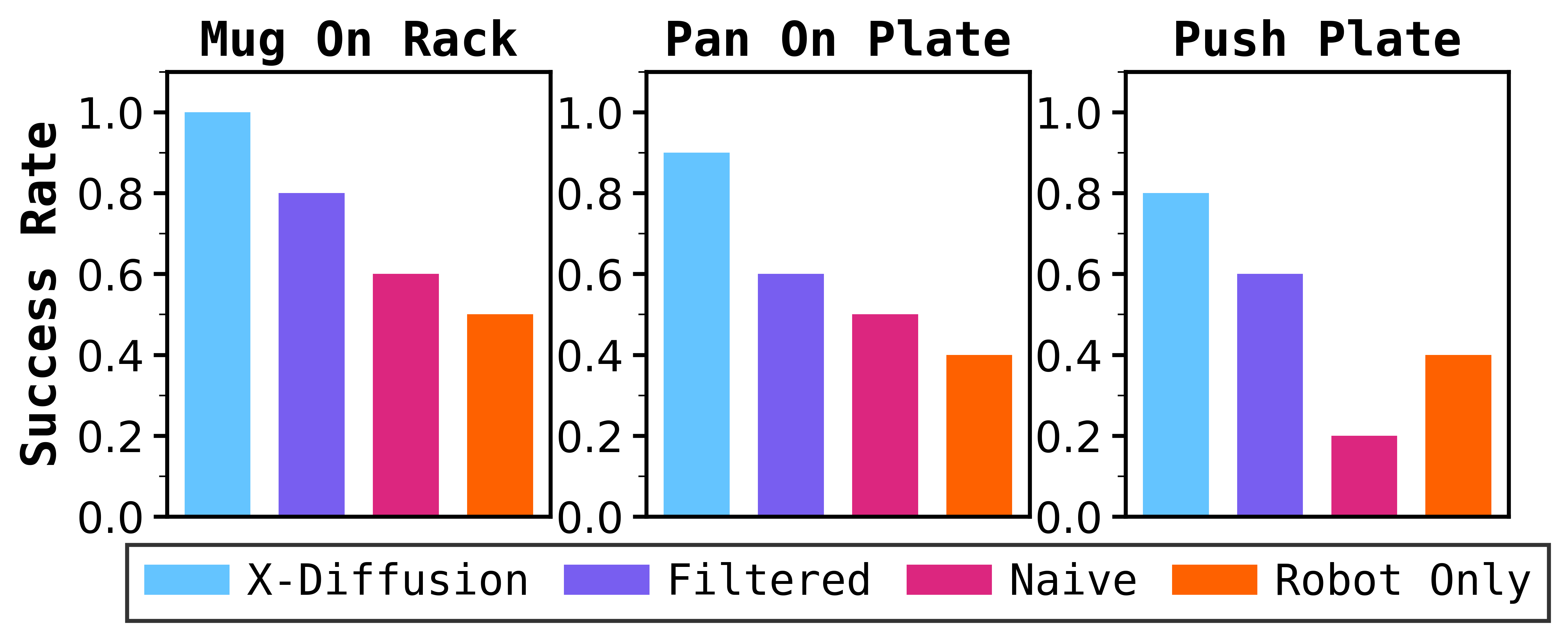}
    \vspace{-4mm}
    \caption{\textbf{Performance vs. Human Data Usage:} We compare \methodname with a policy co-trained on data verified as robot-feasible (\textsc{Filtered}), a naively co-trained policy using all available human data (\textsc{Naive}), and policy trained only on robot data (\textsc{Robot Only}). \methodname consistently outperforms all baselines.}
    \label{fig:oracle}
    \vspace{-2mm}
\end{figure}

\begin{figure}[t!]
    \centering
\includegraphics[width=0.48\textwidth]{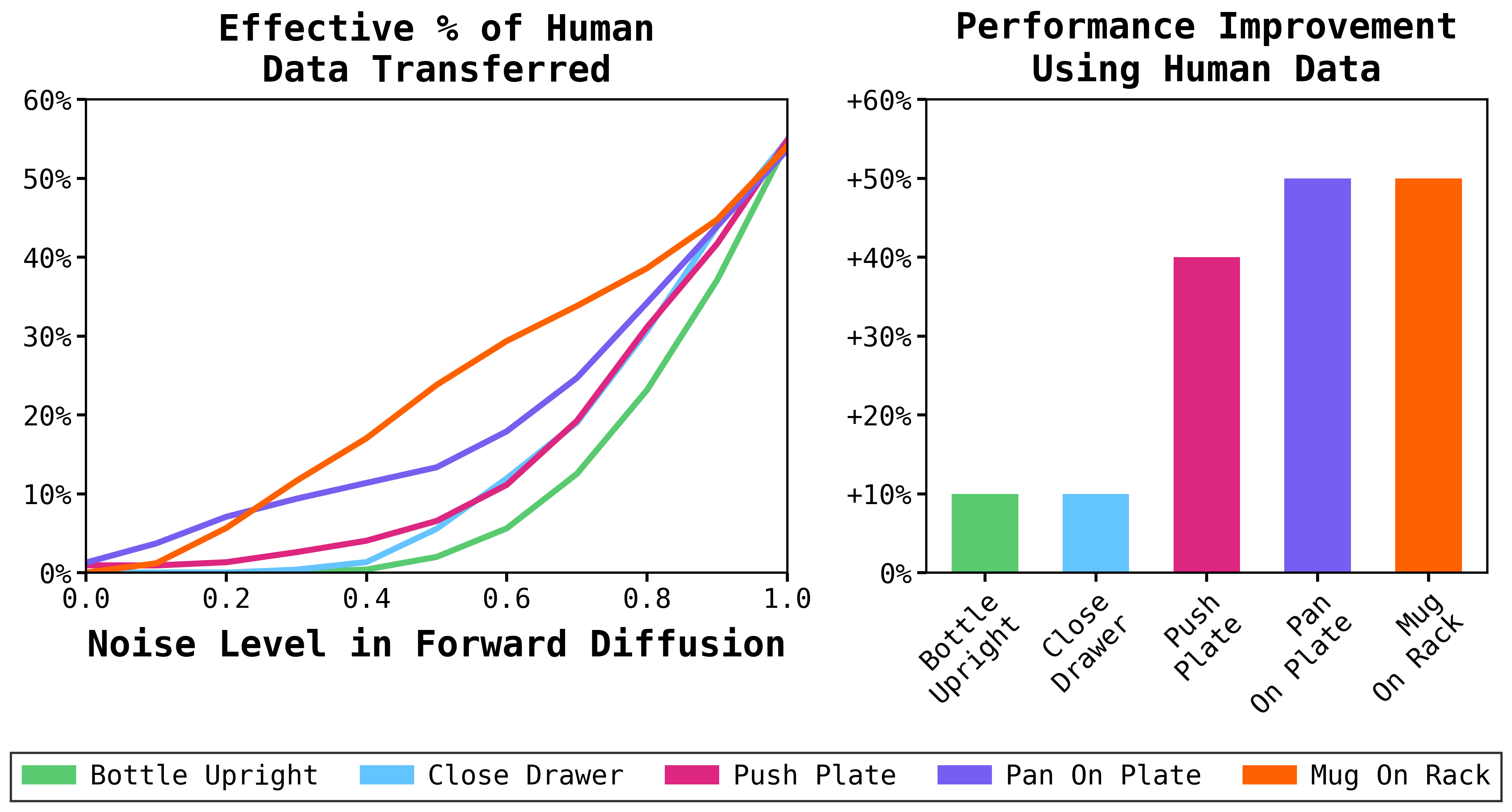}
    \captionsetup{width=0.48\textwidth}
    \vspace{-4mm}
    \caption{\textbf{Quantifying Transfer Learning from Human Data in \methodname:} \textbf{(Left)} For each manipulation task, we measure the fraction of human data incorporated into \methodname\ during training. As the diffusion noise level increases, \methodname\ uses a larger fraction of human data. This fraction varies across tasks; for example, \texttt{Mug On Rack} consistently uses a larger fraction of human data than \texttt{Bottle Upright}. \textbf{(Right)} We measure the performance gain of \methodname\ when trained with human data relative to a baseline trained only on robot data. All tasks benefit from human data, and tasks that incorporate more of it into training, such as \texttt{Mug On Rack}, show larger improvements than tasks that use less, such as \texttt{Bottle Upright}.}
\label{fig:linegraph}
\vspace{-6mm}
\end{figure}


A central question in cross-embodiment learning is whether human demonstrations yield \emph{positive transfer} for robot policy learning, i.e., whether adding human data improves performance relative to training on robot data alone. We find that \methodname\ achieves positive transfer by selectively incorporating human data in a task-dependent manner. Figure~\ref{fig:linegraph} quantifies the amount of transfer across tasks. On the left, we quantify the fraction of human data incorporated into training across different noise levels in the diffusion process. We show that \methodname\ benefits from transfer learning from human data to varying degrees across all five tasks. \texttt{Mug On Rack} and \texttt{Pan On Plate} integrate a larger fraction of human data throughout the diffusion process. \texttt{Bottle Upright} integrates substantially less data, suggesting that its human demonstrations are less dynamically compatible with robot execution. On the right, we quantify \emph{positive transfer} as the performance gain of \methodname\ with human data relative to a robot-only baseline. Across all tasks, incorporating human data improves performance, and tasks that integrate more human data show larger gains. Together, these results show that the benefit of transfer learning from human data is task-dependent. Higher performance gains are observed when the human demonstrations are more aligned with the dynamics of robot execution. 

Importantly, the transfer achieved by \methodname\ is consistently \emph{positive}. In contrast, Fig.~\ref{fig:barplots} shows that prior cross-embodiment baselines often suffer from negative transfer and can perform worse than training on robot data alone. Fig.~\ref{fig:oracle} provides a more systematic ablation by varying different choices of the data used to train \methodname. This shows that the benefit of human supervision depends critically on selecting demonstrations that are truly transferable to the robot. Positive transfer does not arise simply from indiscriminately adding more data, but from selectively incorporating dynamically feasible human actions.

\section{Discussion}

In this paper, we propose \methodname, a cross-embodiment learning framework 
for co-training robot policies on human and robot data. Our key idea is to view dynamically
infeasible cross-embodiment demonstrations as an analog to
low-quality data and leverage recent advances in learning from noisy data \cite{daras2023ambient, daras2023consistent, daras2024consistent, daras2024much, daras2025ambientdiffusionomnitraining} to effectively integrate them into diffusion policy learning.
\methodname\ trains a classifier to identify the minimum noise level where a human action becomes indistinguishable from a robot action, incorporating human actions into training only when they are noised beyond this threshold.
This provides coarse task guidance while avoiding the transfer of physically infeasible behaviors.
This selective co-training enables effective use of human datasets for robot policy learning, allowing \methodname\ to consistently outperform robot-only policies and prior co-training baselines across five manipulation tasks.

\textbf{Limitations.} In our work, we train \methodname on a limited number of robot and human demonstrations in a calibrated multi-camera environment. Future works will attempt to train policies on large-scale datasets and learn from unstructured internet-scale human videos.


\section{Acknowledgments}
The research is partially supported by a gift from Ai2, a NVIDIA Academic Grant, and DARPA TIAMAT program No. HR00112490422. This research is also supported in part by Google Faculty Research Award, OpenAI SuperAlignment Grant, ONR Young Investigator Award, NSF RI \#2312956, and NSF FRR \#2327973. Its contents are solely the responsibility of the authors and do not necessarily represent the official views of DARPA.

\bibliographystyle{IEEEtranBST/IEEEtran}
\bibliography{IEEEtranBST/IEEEabrv,refs_abbr}



\end{document}